\documentclass[letterpaper, 10 pt, conference]{ieeeconf}  

\IEEEoverridecommandlockouts                              

\usepackage{epsfig} 
\usepackage{graphicx} 
\usepackage{mathptmx} 
\usepackage{times} 
\usepackage{amsmath} 
\usepackage{bm}
\usepackage{amssymb}  
\usepackage{booktabs}
\usepackage{tikz}
\usepackage{xcolor}
\usepackage{siunitx}
\usepackage{url}

\usepackage{caption}
\usepackage{subcaption}

\usetikzlibrary{calc}
\usetikzlibrary{decorations.markings}
\usetikzlibrary{shapes,arrows,positioning}
\usetikzlibrary{circuits.ee.IEC}
\usetikzlibrary{calc}
\tikzstyle{block} = [draw, fill=blue!20, rectangle, 
    minimum height=3em, minimum width=6em]
\tikzstyle{sum} = [draw, fill=blue!20, circle, node distance=1cm]
\tikzstyle{input} = [coordinate]
\tikzstyle{output} = [coordinate]
\tikzstyle{pinstyle} = [pin edge={to-,thin,black}]

\title{\LARGE \bf
 Precise Robot Localization in Architectural 3D Plans
}

\author{Hermann Blum\authorrefmark{1}, Julian Stiefel\authorrefmark{1}\authorrefmark{2}, Cesar Cadena\authorrefmark{1}, Roland Siegwart\authorrefmark{1} and Abel Gawel\authorrefmark{1} 
\thanks{\authorrefmark{1} Autonomous Systems Lab, ETH Zurich}
\thanks{\authorrefmark{2} Multi Scale Robotics Lab, ETH Zurich}
\thanks{{\tt\small\{blumh, jstiefel, cesarc, rsiegwart, gawela\}@ethz.ch}}%
\thanks{This work was partially supported by the Swiss National Science Foundation  (SNF),  within  the  National  Centre  of  Competence  in  Research  on Digital Fabrication and by the HILTI group.
}}

\begin{document}

\begin{minipage}{\textwidth}
\copyright 2020 IEEE. Personal use of this material is permitted. Permission from IEEE must be obtained for all
other uses, in any current or future media, including reprinting/republishing this material for advertising
or promotional purposes, creating new collective works, for resale or redistribution to servers or lists, or
reuse of any copyrighted component of this work in other works.\\

\end{minipage}

\maketitle
\thispagestyle{empty}
\pagestyle{empty}

\begin{abstract}

This paper presents a localization system for mobile robots enabling precise localization in inaccurate building models. The approach leverages local referencing to counteract inherent deviations between as-planned and as-built data for locally accurate registration. We further fuse a novel image-based robust outlier detector with LiDAR data to reject a wide range of outlier measurements from clutter, dynamic objects, and sensor failures. We evaluate the proposed approach on a mobile robot in a challenging real world building construction site. It consistently outperforms the traditional ICP-based alingment, reducing localization error by at least 30\%.

\end{abstract}

\section{Introduction}

The building construction industry is amongst the least digitized, and most dangerous industries for  workers with stagnating productivity.
Industry leaders, worker associations, and changing regulations push the industry towards large innovation needs.
One emerging technology are assistive mobile robots in building construction enabling both higher degrees of digitized processes, and reduced risk for human workers~\cite{ardiny2015autonomous}.
To this end, mobile robots perceive the environment with digital sensors, offering ease of relating information from digital building models to robot perception.
Localizing robots in these building models with high accuracies then enables them to perform building tasks with respect to these data, or accurately track construction progress. 

Conventional methods rely on using sensing, e.g., total stations, or augmentation of the environment, e.g., with artificial markers, to achieve high accuracies. However, these solutions require line-of-sight to a manually placed total station or marker and therefore depend on time-consuming manual preparation for every site, thus limiting the ease and autonomy of such systems. Furthermore, accurately localizing model information with mobile robots is not straightforward, due to the following challenges: 
\begin{enumerate}
\item Multi modality: Robotic sensor data is recorded with extereoceptive sensors, e.g., cameras and LiDARS while building models are typically manually created 3D mesh models, rendering data integration and registration between these domains difficult.
\item Deviations: As-built environments typically deviate from the as-planned state by up to several centimeters (\SI{}{cm}) rendering global references infeasible for accurate task execution.
\item Clutter: Real construction sites contain numerous temporary artefacts that are not modelled in the building models, as well as dynamic actors, such as human workers.
\end{enumerate}

\begin{figure}[t]
    \centering
    \includegraphics[width=\linewidth]{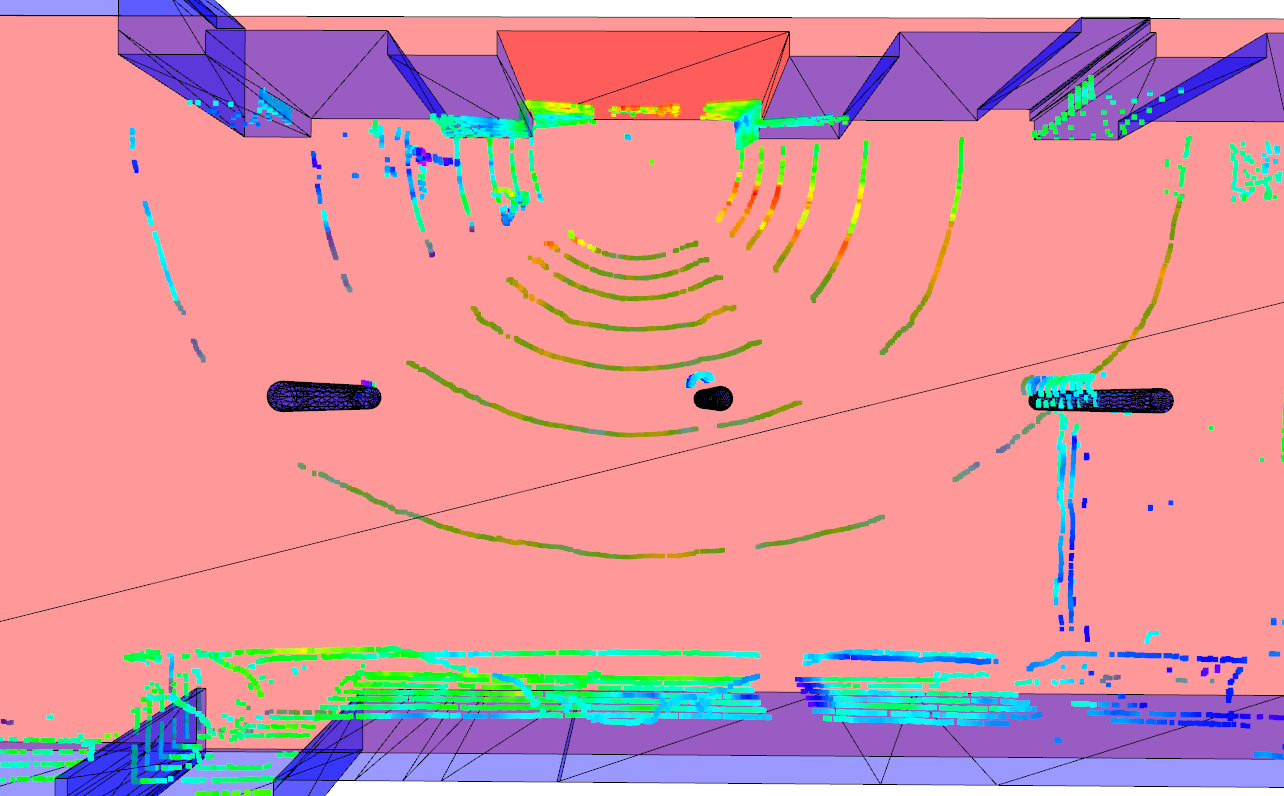}
    \caption{Our proposed method localizes a LiDAR scan against reference surfaces (red) in the mesh of a building model. The scan of the LiDAR is fused with semantic information and points are colored according to their high (red) or low (blue) probability of belonging to building structure. The visible mismatches between the LiDAR scan and the columns are not a localization failure, but highlight the challenge of localizing within a built structure deviating from a planned model.}
    \label{fig:teaser}
\vspace{-0.5cm}
\end{figure}

In this work, we propose a robotic localization system that addresses these challenges by combining locally referenced sensing with a learning-based sensor fusion approach for robust outlier rejection. The system therefore requires only on-board sensing without any artificial preparation of the site. We address the multi-modality aspect by converting the mesh model into a sparse point-cloud which is easily referenced to sensor point-cloud data using standard scan-matching algorithms. Furthermore, we propose a task-based referencing solution, yielding locally accurate localization. Finally, we fuse a novel learning-based robust detector for outlier rejection on image data with LiDAR data, able to reject clutter that is outside the training distribution. Our system is tested on a mobile robot in realistic construction environments, showing a reduction of localization error of at least 30\%.
In summary, this paper presents the following contributions:
\begin{enumerate}
\item Fusing range sensing data with learned robust outlier filters, producing semantically annotated sensor data that can be leveraged in 3D architectural floorplan localization.
\item An error metric and local referencing system enabling improved localization accuracies for locally referenced building tasks.
\item Evaluation of the proposed methods and the traditional ICP baseline with respect to the aforementioned challenges.
\end{enumerate}

This paper is structured as follows: In Section~\ref{sec:related_work}, we present related works on robotic floorplan localization and semantic localization. In Section~\ref{sec:method}, we  present our proposed 3D architectural floorplan method, and report experimental results on a mobile robotic platform in Section~\ref{sec:experiments}. Finally, we discuss and conclude our findings in Sections~\ref{sec:discussion} and~\ref{sec:conclusion} respectively.

\pgfdeclarelayer{background}
\pgfdeclarelayer{foreground}
\pgfsetlayers{background,main,foreground}
\begin{figure*}
\centering

\begin{tikzpicture}[every node/.style = {outer sep=3pt, inner sep=0}, line width=0.7pt]
  \node[label=-90:{RGB Image\strut}] (rgb) {\includegraphics[height=1.8cm]{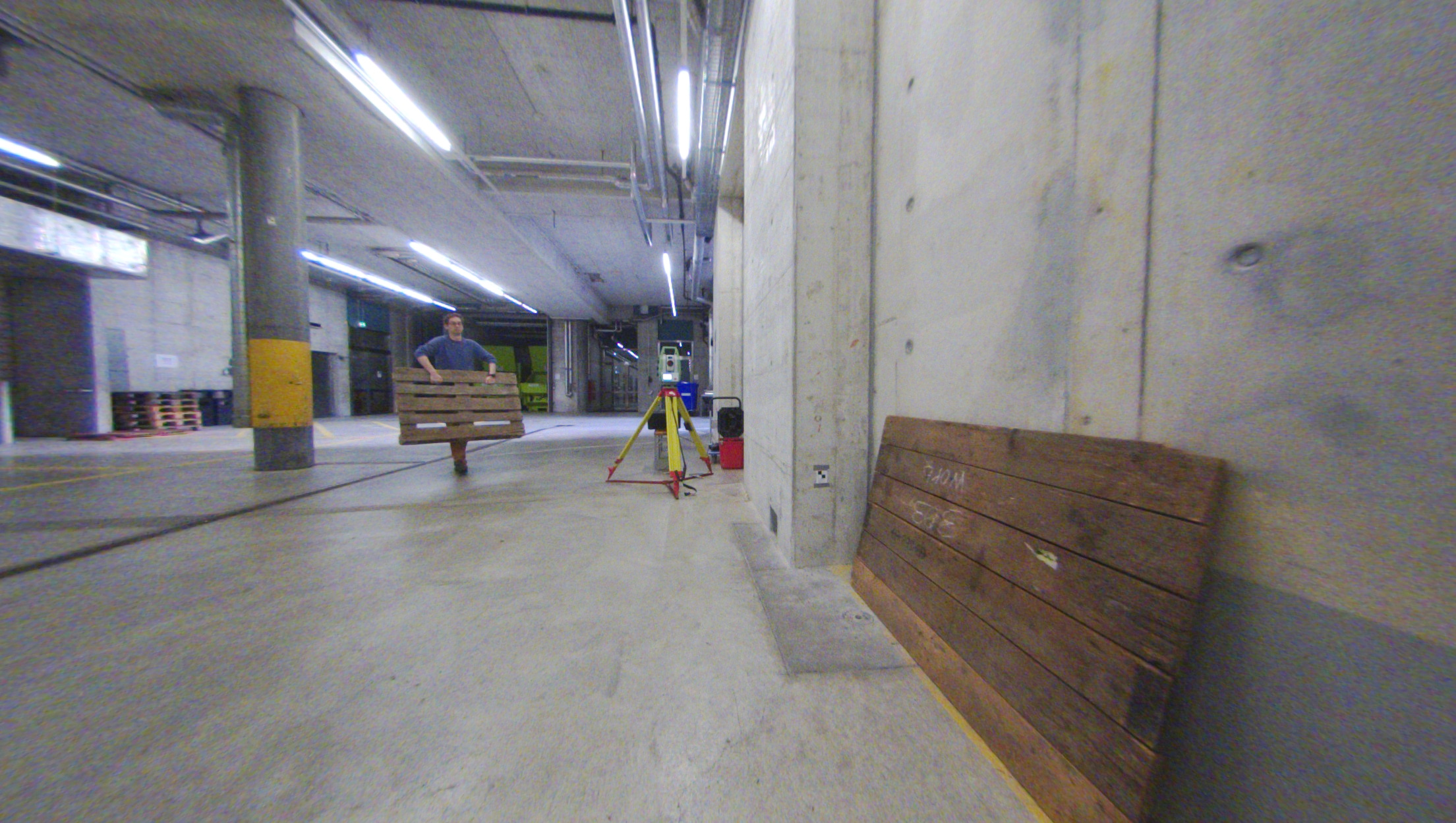}};
  \node[right = 0.45 cm of rgb, label=-90:{Segmentation\strut}] (segmentation) {\includegraphics[height=1.8cm]{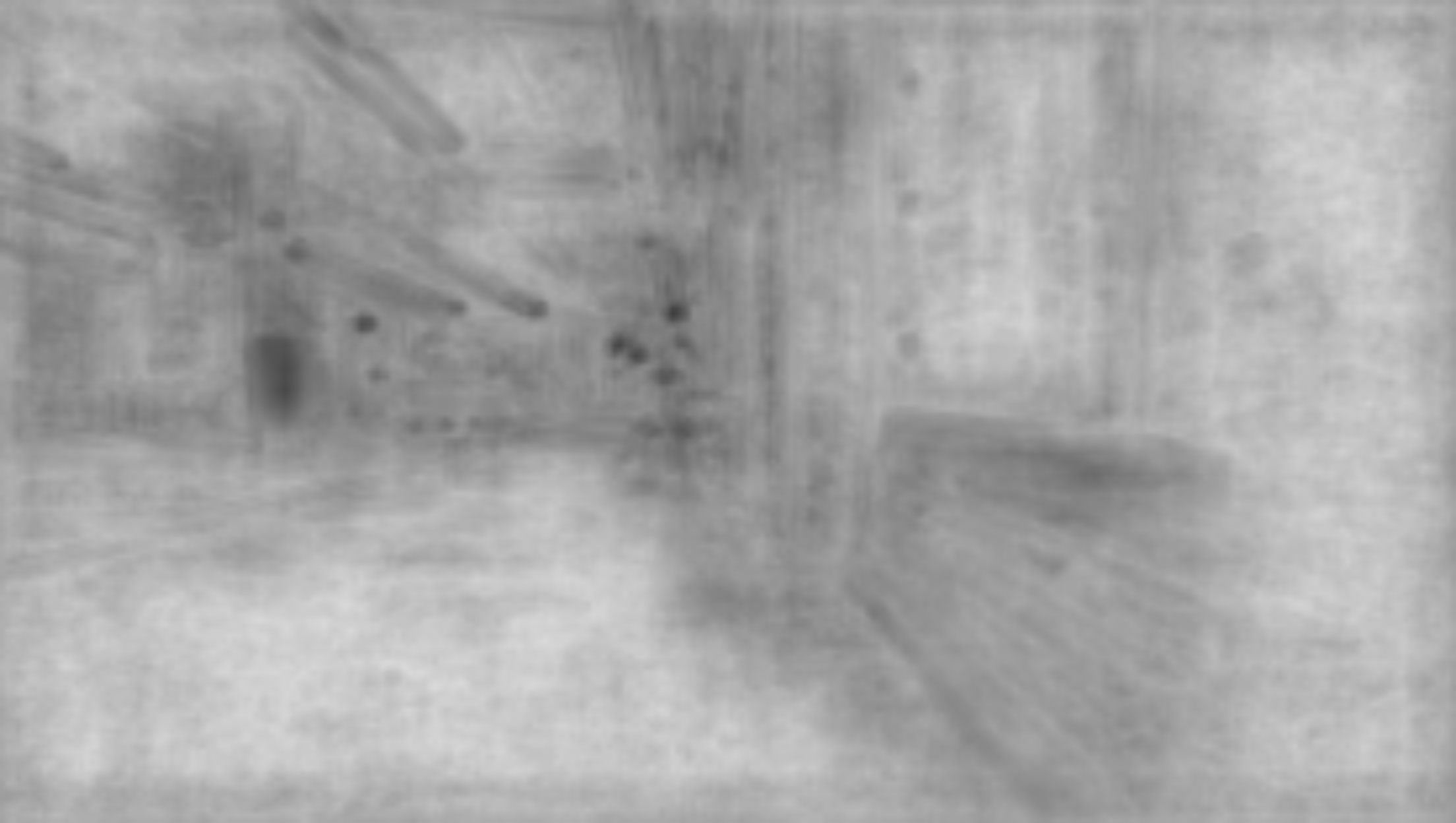}};
  \node[right = 0.45 cm of segmentation, label=-90:{Fusion into Scan\strut}] (scan) {
  \includegraphics[height=1.8cm]{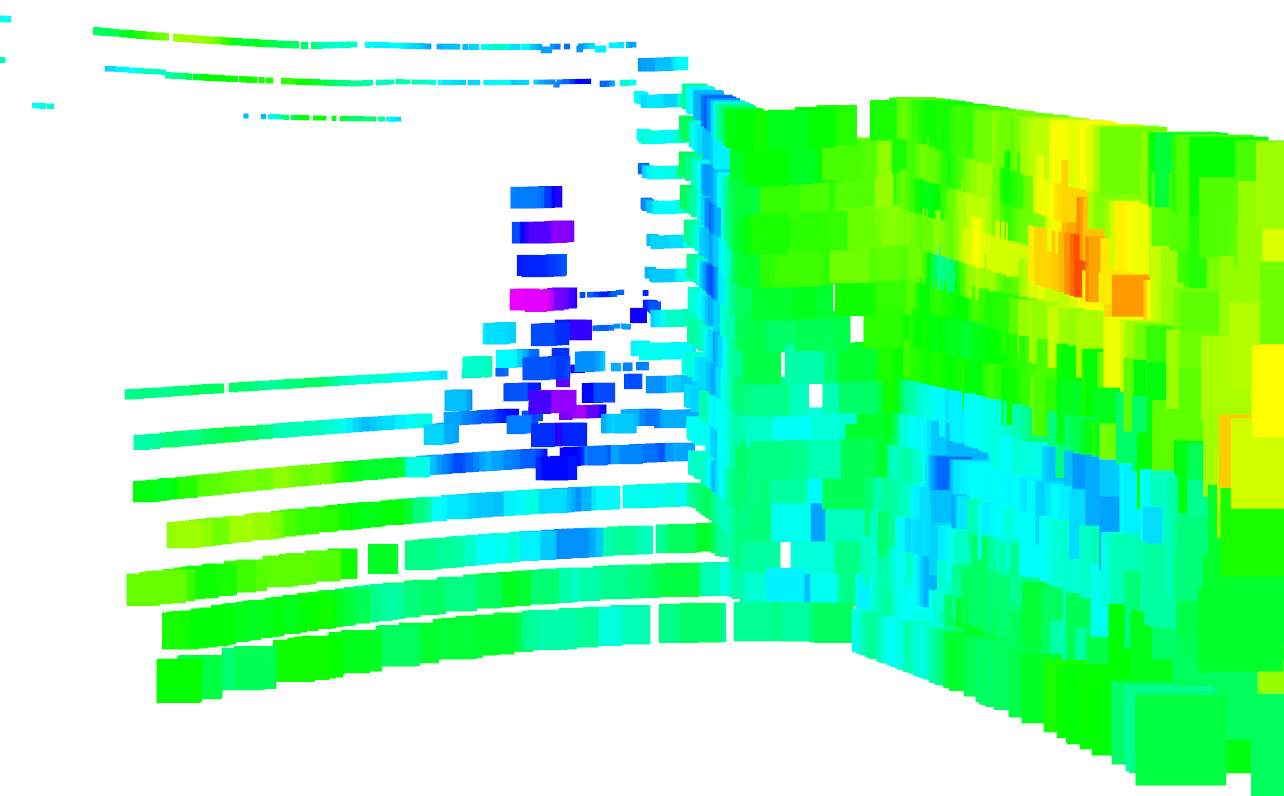}
  };
  \node[right = 0.45 cm of scan, label=-90:{Alignment\strut}] (localisation) {\fcolorbox{red}{red}{\includegraphics[height=1.8cm]{figures/localisation_screenshot.png}}};
  \node[right = 0.45 cm of localisation, label=-90:{References in Model}] (model) {\includegraphics[height=1.8cm]{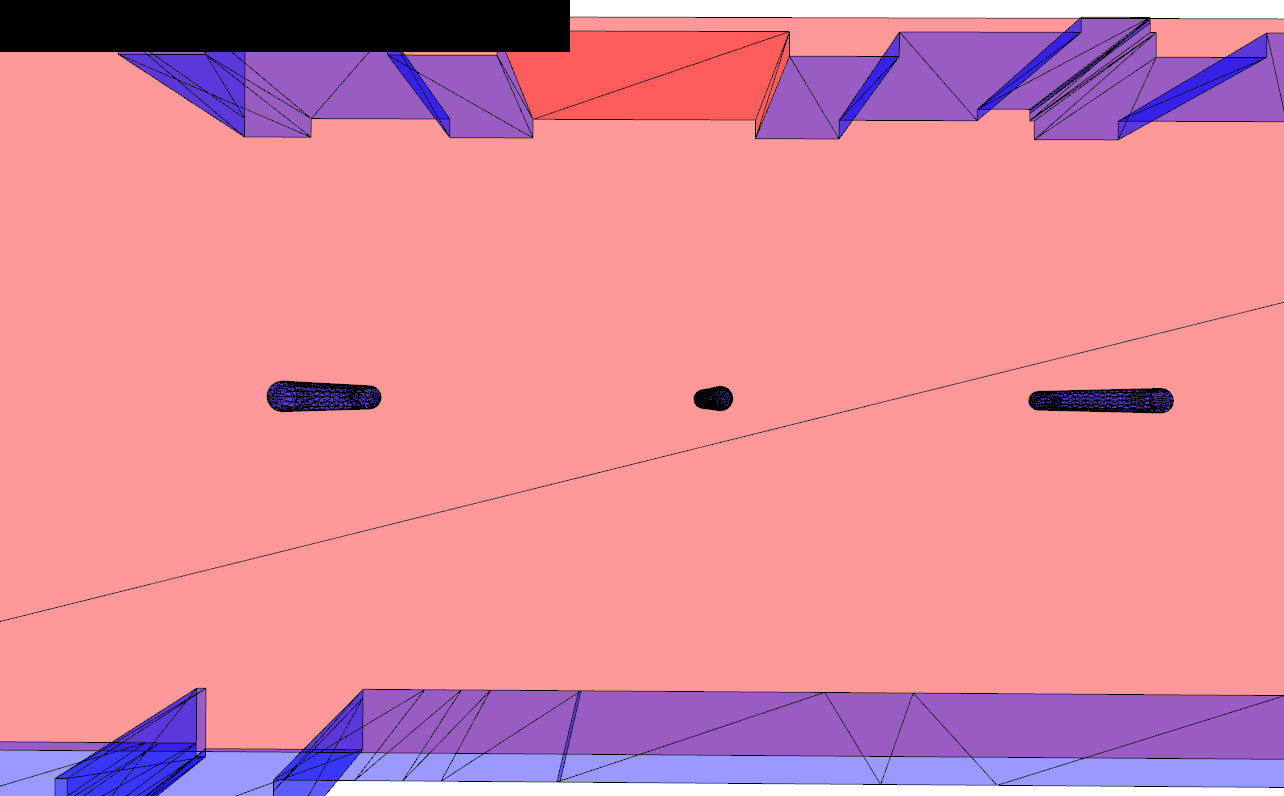}};
  
  \draw [->] (rgb) -- (segmentation);
    \draw [->] (segmentation) -- (scan);
    \draw [->] (scan) -- (localisation);
    \draw [->] (model) -- (localisation);
\end{tikzpicture}

\caption{Overview of the proposed method: RGB images are segmented into foreground and background using a robust segmentation network~\cite{Marchal2019-dv}. These pixelwise scores are then propapgated and fused with LiDAR scans. The classifications are then used as weights in the floorplan localization which performs a weighted selective localization in a 3D floorplan given local task references.}
\label{fig:overview}
\vspace{-0.5cm}
\end{figure*}

\section{Related Work}
\label{sec:related_work}
\subsection{Robot Localization in Architectural Plans}
While registration of scans to building models is a well-studied problem in surveying~\cite{tam2012registration}, the application was studied less extensively in robotics. With the raise of LiDAR sensors, different works studied 2D localization within floorplans~\cite{Boniardi2017-dp, Hess2016-lx}. However, to execute construction tasks, 2D localization is insufficient. Walls might be tilted or floors elevated such that water can run off. At higher levels of accuracy, the flat and rectangular world assumption therefore is no longer valid. \cite{Gawel_undated-qs}~therefore studied the extension of conventional methods to 3D, but rely on a special measurement system to localize their robot's endeffector with respect to local reference walls. In this work, we study localization methods that do not require manipulators mounted on the robot. Bosch{\'e}~\cite{bosche2010automated} shows ICP-based localization of 3D scan data within building mesh models under the assumption of negligible deviations between model and reality.

In case that LiDAR sensors are not available, research has made progress to extract floorplan-like information out of camera images~\cite{Phalak2019-uq}. Boniardi et al.~\cite{boniardi2019robot} studied how such information can be used to localize a robot in 2D within a given floorplan. Mendez at al.~\cite{Mendez2017-jj} localize an RGB camera in floorplans based on semantic information and can show that their method does not benefit from range information.

\subsection{Semantically Enriched Localization with LiDARs}
\cite{segmap2018} show that semantic cues can robustify global localization techniques. However, the considered semantics only considered distinct classes as information cue, while our approach is more fine-grained using semantic information for filtering and weighting measurements.
Closest to our work is~\cite{chen2019suma++}. The authors propose to use a semantic segmentation network on LiDAR data and a semantic consistency term in a surfel-based map representation to filter dynamic objects over multiple observations. Furthermore, the work proposes to weigh associations of an ICP registration using semantic label classification scores. Our work differs in the used network that does not require pre-knowledge about semantic classes, and can generalise to reliably detecting outliers outside its training distribution. 

\subsection{LiDAR Registration}
Among the first papers, \cite{besl1992method} describe the use of ICP for registration of 3D shapes, i.e., an iterative error minimization over 3D point-correspondences. A multitude of variants were proposed since, including more generalized formulations also including probabilistic measurements~\cite{segal2009generalized}, and variants that skip the re-association step~\cite{zhou2016fast}. A comprehensive overview is given in~\cite{pomerleau2013comparing}. Further notable registration algorithms rely on feature extraction~\cite{zhang2014loam}, or representing data as a probability density~\cite{biber2003normal}.
\cite{sandy2016autonomous} demonstrate ICP for localization with respect to a known object in the environment using a LiDAR scanner, achieving sub-centimeter accuracies.
\cite{rowekamper2012position} achieve sub-\SI{}{cm} accuracies using 2D LiDAR localization in a reference scan of the environment. However, the authors also introduce clutter and dynamics to the scenes which heavily degraded the achievable accuracies. Without modifying the principal ICP solution, our method proposes to use only partial data, i.e., local references for scan registration. Furthermore, we explicitly consider outliers. The method weighs or filters associations between scans and building model by fusing scan points with the output of a image-based CNN~\cite{Marchal2019-dv} that quantifies the likelihood of measurements being outliers.

\section{Method}
\label{sec:method}
An overview about the proposed method can be found in Figure~\ref{fig:overview}.
In the following, we first describe the input processing pipeline for LiDAR scans and define the general problem of registering and aligning scans to building models. Subsequently, we describe how further information on the task references can be used to define the alignment more accurately. The whole method assumes a good initial guess for the robot pose to coarsely align the building plan with the initial pose of the robot. This initialization can be provided manually or with a global localization method.

\subsection{Semantic Filtering}
\label{subs:method_semantic}
Classic semantic segmentation of images or pointclouds relies on large datasets to train networks to reasonable performace. Unfortunately, there are no such labelled datasets available for construction environments. Moreover, construction environments are highly dynamic with a lot of potential object classes appearing, whereas we are only interested in background-foreground segmentation, i.e., separating building surfaces from any other scene contents. Due to the limits of available datasets and the potential problems with training on a fixed set of classes, we use an alternative segmentation method described in~\cite{Marchal2019-dv}. The density estimation network from Marchal et al. is trained on the NYU Indoor Room Dataset~\cite{silberman2012indoor}, but able to generalize better than classical methods to new environments. In particular, we use their best performing model, which is a regression over density estimation at multiple layers of the feature extractor. Instead of a binary segmentation, the method returns a per-pixel score that is higher if the pixel belongs to background structure of a building. We will hereafter refer to this score as density value $d$.

The proposed localization system relies on a multi-camera plus LiDAR setup with known intrinsic, extrinsic and timestamp calibrations. Whenever a new LiDAR scan arrives, we rectify the corresponding set of images $I = \lbrace I_0, I_1, ...\rbrace$ and process them in the density estimation network~\cite{Marchal2019-dv}.\\
From the scanned LiDAR points $P = \lbrace p_1, p_2, ...\rbrace$, we project each point onto each image plane. For those points $p_i$ within the field of view of a camera, we assign the density value $d_i$ to the point. We reject all points that cannot be projected onto any of the images, but the amount of such points is negligibly small due to the high FoV of the camera lenses.

To localize within the building model, the scan $P$ is aligned to the model with point-to-plane ICP: 
\begin{align*}
    T_\textrm{icp} = ICP(P, S) &= \arg \min \limits_T \sum_i w_i \, c\left\lbrack (p_i - m_i) \cdot \bm{n}(m_i) \right\rbrack
\end{align*}
where $c()$ is a cost function, $m_i$ is a matched point of $p_i$ in the building model $S$ and $\bm{n}(m_i)$ is the surface normal at that point.

For the weight $w$, we propose two variants:

\begin{align}
    w_i &= \left\lbrace \begin{array}{cl} 0 & d_i < \delta \\ 1 & d_i \geq \delta \end{array}  \right. \label{eq:binary}\\
    w_i &= \max(0, a \, d_i - \delta') \label{eq:weighting}
\end{align}
Where $a$ is a normalization factor such that $\max \limits_i w_i = 1$.
Equation (\ref{eq:binary}) corresponds to a binary segmentation of the scan where only those points which belong to background structure are considered in the ICP problem and equation  (\ref{eq:weighting}) assigns higher weights to points that are more likely to belong to background structure. The non-binary weighting does not rely on perfect predictions from the density estimation network and instead uses the prediction confidence in the ICP optimization.

\subsection{(Selective) Localization}
\label{subs:method_selective}

\begin{figure}
    \centering
    \begin{subfigure}[b]{0.4\linewidth}
         \centering
         \includegraphics[width=\textwidth]{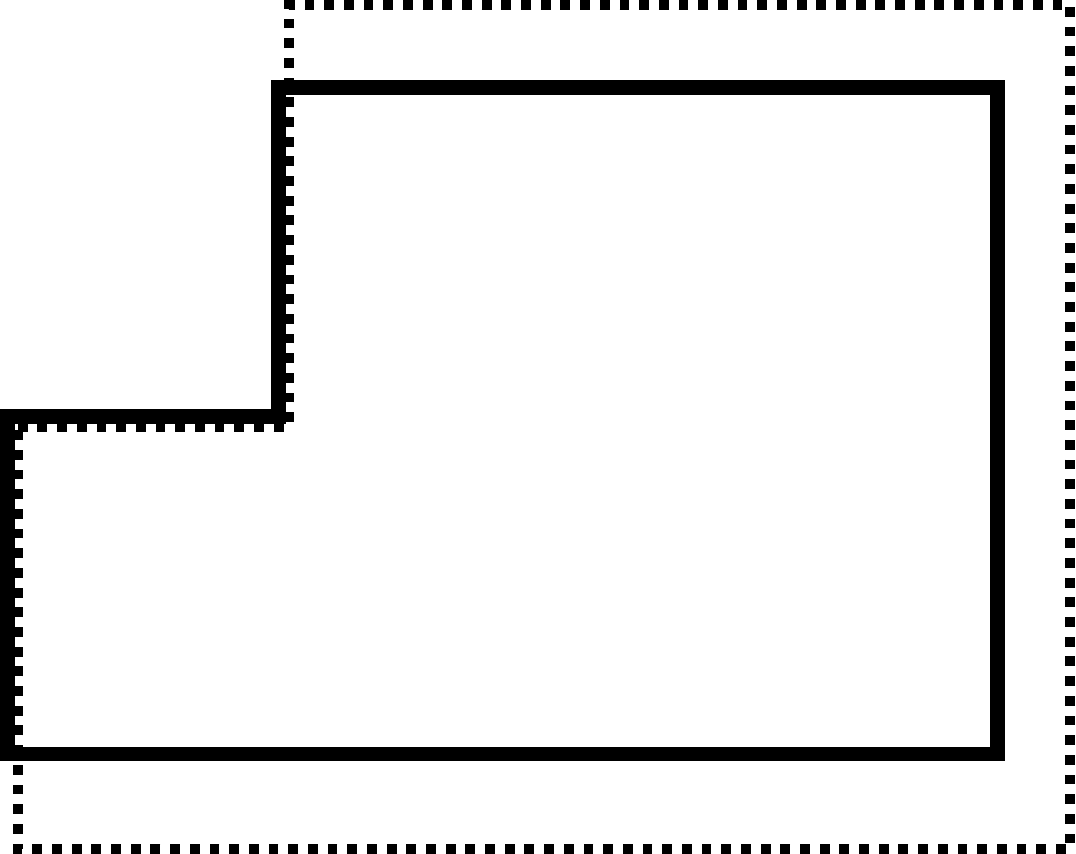}
         \caption{ambiguours alignment}
     \end{subfigure}
     \hfill
     \begin{subfigure}[b]{0.4\linewidth}
         \centering
         \includegraphics[width=\textwidth]{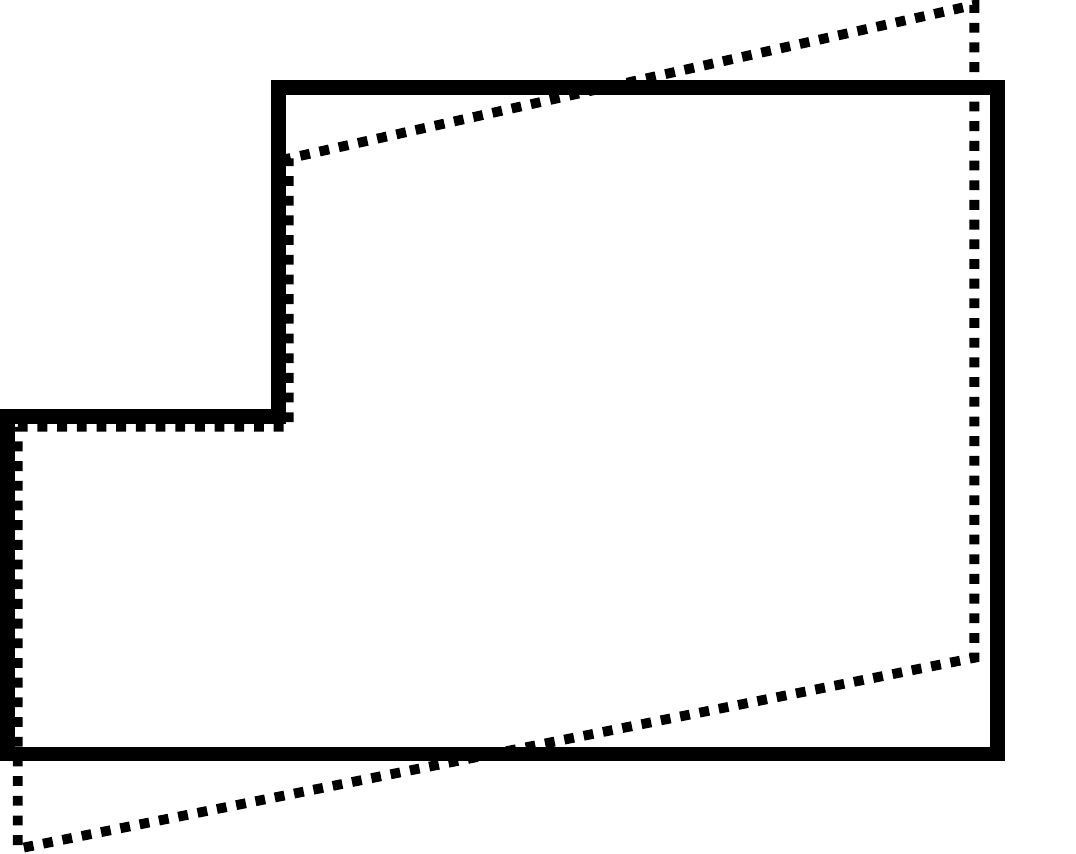}
         \caption{high-error alignment}
     \end{subfigure}
    \caption{Illustrations for deviations between as-planned (solid) building model and as-built scan from the robot (dotted). In (a), the alignment of the horizontal walls is ambiguous and traditional ICP will align that wall with more points in the scan. In (b), any alignment will yield high errors and the solution in general depends on the settings for outlier filtering.}
    \label{fig:alignment_issues}
\vspace{-0.5cm}
\end{figure}

We define the mesh of the building model as a collection of closed surfaces $S = \lbrace s_0, s_1, ...\rbrace$. In general, we can then localize the robot with $ICP(S, P)$. However, building models are usually far from perfect maps of the actual environment, where during the construction phase walls might be missing, temporary structure such as scaffolding is not mapped and the relative positions of the final walls can deviate from the building model. Some of the typical deviations are illustrated in Figure~\ref{fig:alignment_issues}.

To enable the robot to localize precisely in a building that deviates from the map, we define a subset of walls or surfaces $R \subset S$ with $\exists \; r_1, r_2, r_3: r_1 \nparallel r_2, r_2 \nparallel r_3, r_1 \nparallel r_3, r_1, r_2, r_3 \in R$. The reference surfaces therefore  define a (minimally) sufficient alignment problem for the robot to localize itself, but exclude all surfaces in $S$ that are not relevant to a local task and would in case of deviations disturb the alignment. This is further motivated by most construction tasks being expressed in local reference frames. Within the case of approximately rectangular building structures that we study in our experiments, $R$ is  often the set of surfaces defining a corner in a room.

Initial experiments with our methods showed that in cases where the surfaces in $R$ are only measured with a few points of the LiDAR scan $P$, the ICP alignment can fail with huge errors. We therefore follow these steps:

\begin{enumerate}
    \item Find the alignment\\$T_\textrm{full icp}^{(t)} = ICP(S, P | T^{(t - 1)})$
    \item Refine the alignment to the references\\$T_\textrm{selective icp}^{(t)} = ICP(R, P | T_\textrm{full icp}^{(t)})$
    \item Reject  $T_\textrm{selective icp}^{(t)}$  if $\left| T_\textrm{selective icp}^{(t)} - T_\textrm{full icp}^{(t)} \right|$ is too large.
\end{enumerate}

\section{Experiments}
\label{sec:experiments}
\subsection{Sensor Setup}

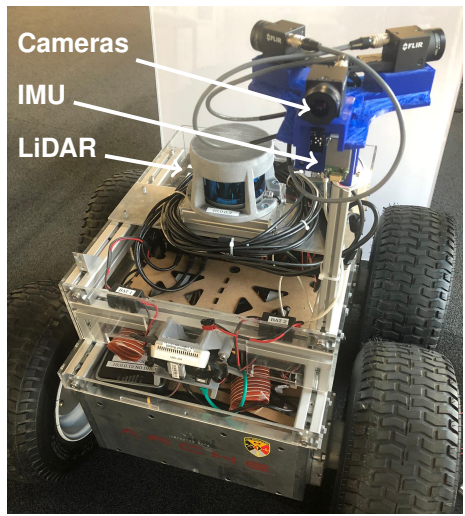
\begin{figure}
    \centering
    \begin{tikzpicture}
    \color{white}
    \sffamily
    \node[inner sep=0] at (0,0) {\includegraphics[width=0.7\linewidth]{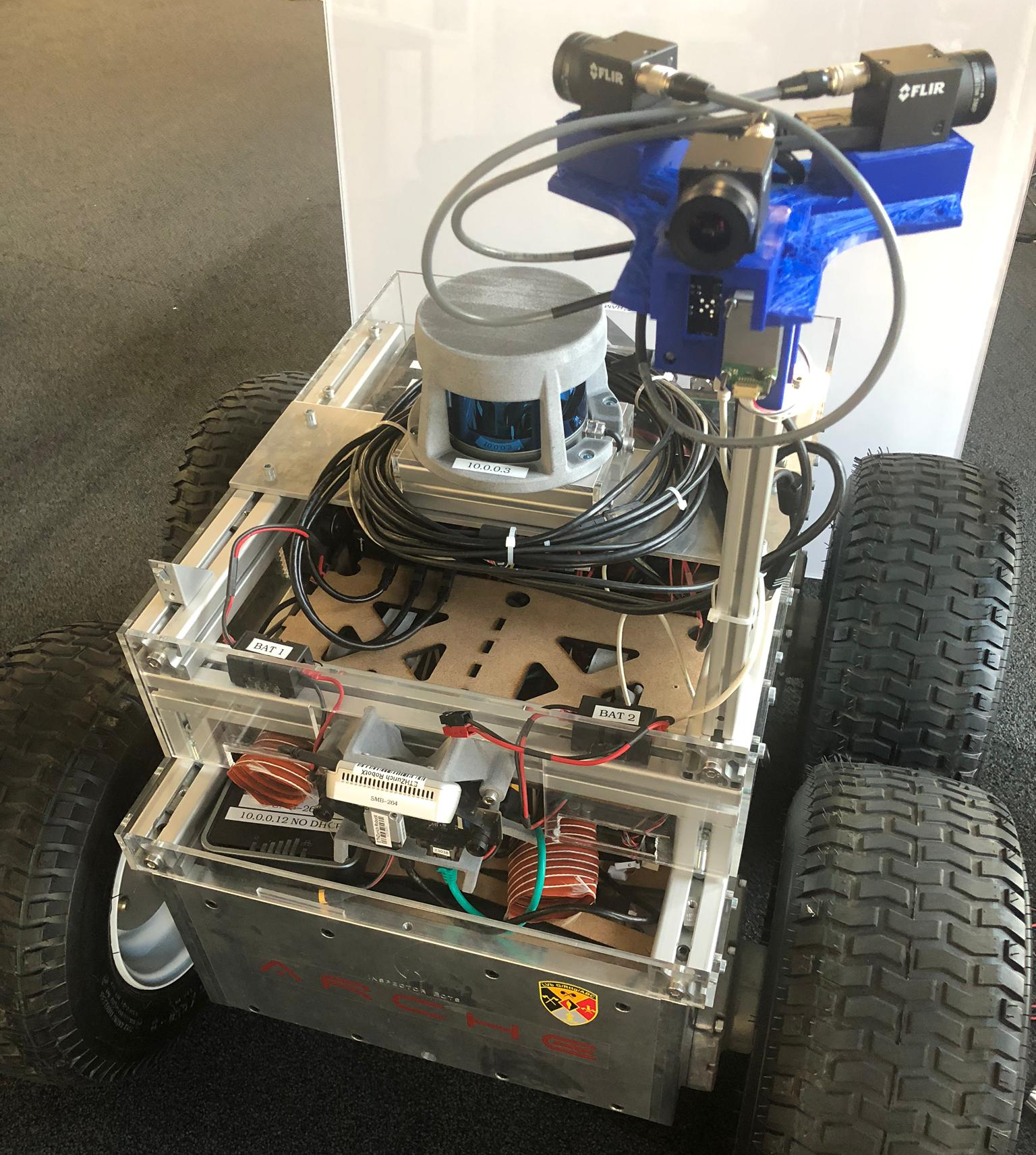}};
    \path (-0.7,1.2) coordinate (lidar)
          (1,2) coordinate (cameras)
          (1.1,1.3) coordinate (imu);
    \node[inner sep=0, anchor=west, outer sep=3pt] (lidarlabel) at (-3, 1.5) {\textbf{LiDAR}};
    \node[inner sep=0, anchor=west, outer sep=3pt] (camlabel) at (-3, 2.9) {\textbf{Cameras}};
    \node[inner sep=0, anchor=west, outer sep=3pt] (imulabel) at (-3, 2.2) {\textbf{IMU}};
    \path[->,white,line width=1.5pt] (lidarlabel) edge (lidar);
    \path[->,white,line width=1.5pt] (camlabel) edge (cameras);
    \path[->,white,line width=1.5pt] (imulabel) edge (imu);
\end{tikzpicture}
    \caption{Sensor setup on the robot.}
    \label{fig:sensor_setup}
\vspace{-0.5cm}
\end{figure}

We conduct our experiments on the \emph{supermegabot}\footnote{\url{github.com/ethz-asl/eth-supermegabot}} platform with a LiDAR and three cameras. The robot is further equipped with an IMU for smooth state estimation. The exact sensor configuration is shown in Figure~\ref{fig:sensor_setup}.

We calibrate all cameras individually using Kalibr\footnote{\url{github.com/ethz-asl/kalibr}}~\cite{Furgale2013-jo, Maye2013-tp} to obtain intrinsics and extrinsics with respect to the IMU. We then record a joined motion of LiDAR and cameras to find their respective transformations using visual-intertial odometry \cite{Schneider2018-gk} with one of the cameras and optimizing the alignment of LiDAR scans with respect to that trajectory\footnote{\url{github.com/ethz-asl/lidar-align}}. Time-synchonization is achieved with the VersaVIS~\cite{tschopp2019versavis} camera trigger board that synchronizes time between the host, all cameras and the IMU, while negligible time-offset is assumed between the host and the LiDAR.

To measure ground-truth, we attach a prism to the robot and measure the position with a total station referenced to the origin of the building plan that the robot is using as a map. We calibrate the prism position with respect to the robot frame by aligning trajectories of the robot moving around. In case that as-built reference walls deviate from the building plan in absolute coordinates, we measure this deviation and correct the ground truth position accordingly.

\subsection{Building Model}
We supply the robot with a 3D mesh of the building that is generated from the available 2D floorplan. Because there are no information of the floor or the ceiling structure available, we add a planar floor to the mesh and give all walls the same height. Reference surfaces are specified as a subset of the same mesh.

\subsection{Stationary Robot}
\begin{figure}
    \centering
    \begin{subfigure}[b]{0.8\linewidth}
         \centering
         \includegraphics[width=0.3\textwidth]{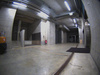}
         \fcolorbox{red}{red}{\includegraphics[width=0.3\textwidth]{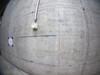}}
         \includegraphics[width=0.3\textwidth]{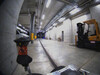}
         \caption{Location A}
     \end{subfigure}
    \begin{subfigure}[b]{0.8\linewidth}
         \centering
         \includegraphics[width=0.3\textwidth]{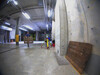}
         \fcolorbox{red}{red}{\includegraphics[width=0.3\textwidth]{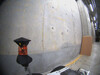}}
         \includegraphics[width=0.3\textwidth]{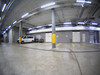}
         \caption{Location B}
     \end{subfigure}
    \begin{subfigure}[b]{0.8\linewidth}
         \centering
         \includegraphics[width=0.3\textwidth]{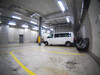}
         \fcolorbox{red}{red}{\includegraphics[width=0.3\textwidth]{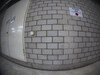}}
         \includegraphics[width=0.3\textwidth]{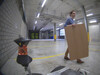}
         \caption{Location C}
     \end{subfigure}
    \begin{subfigure}[b]{\columnwidth}
        \includegraphics[width=\columnwidth]{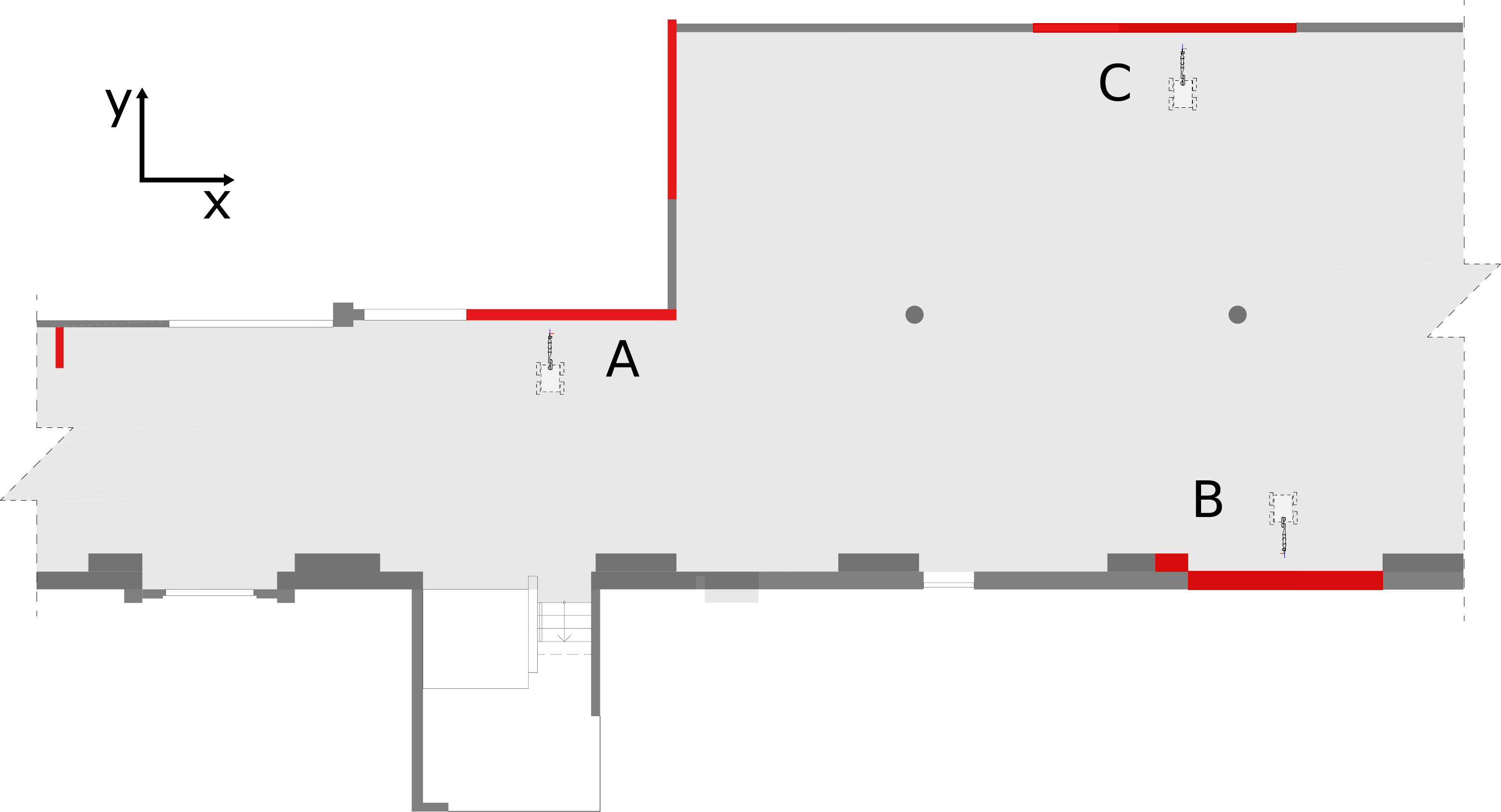}
        \caption{Site Overview}
    \end{subfigure}
    \caption{Selected locations and reference walls (red) for the stationary robot experiment. All locations also have the floor as a reference surface. The red border around the images corresponds to the wall-facing camera.}
    \label{fig:building_plan}
\end{figure}
We conduct experiments with a stationary robot while a construction worker is moving around the robot and clutter objects such as wooden boards and equipment boxes are placed to partially obstruct measurements to reference walls. For each location, we analyze a sequence of around 1 min. or 300 LiDAR scans.

\begin{table}[]
    \centering
    \begin{tabular}{ccc}
        \toprule
        clutter & Loc. A [mm] & Loc. B [mm] \\
        \midrule
        no & 157 & 74\\
        yes & 218 & 83\\
        \bottomrule
    \end{tabular}
    \caption{Influence of clutter on the localization accuracy (rmse in mm). For this comparison, the robot localizes the full LiDAR scan in the full building model.}
    \label{tab:clutter}
\vspace{-0.5cm}
\end{table}

To verify the influence of clutter objects and moving workers on the general localization performance of a robotic system, we compare the situations with and without clutter in Table~\ref{tab:clutter}. The clutter around location C was fixed to the environment, therefore the comparison could only be conducted for locations A and B.

To simulate a severe deviation, we subsequently move the upper and lower structure in Figure~\ref{fig:building_plan} apart by 0.3m in the mesh supplied to the robot.
We compare all combinations of methods described in section~\ref{sec:method} to a baseline of ICP alignment between building model and LiDAR scan. We measure repeatability by the trace and maximum eigenvalue of the covariance matrices for position and rotation. For accuracy, we compare the estimated position of the leica prism given the robot's pose and the measured position. This measurement therefore relies on the estimation of the full robot pose. The results are given in tables~\ref{tab:stationary_a}, \ref{tab:stationary_b} and \ref{tab:stationary_c} for the three different test locations respectively. Additionally, we report the fraction of LiDAR scans that could not be localized either because ICP did not converge or the solution was rejected (see section~\ref{subs:method_selective}). To account for non-determinisms in the localization pipeline, all values are averages over three executions on the same input data.

\begin{figure*}
    \centering
    \begin{subfigure}[b]{\linewidth}
    \includegraphics[width=\linewidth, trim=0 350 0 400, clip]{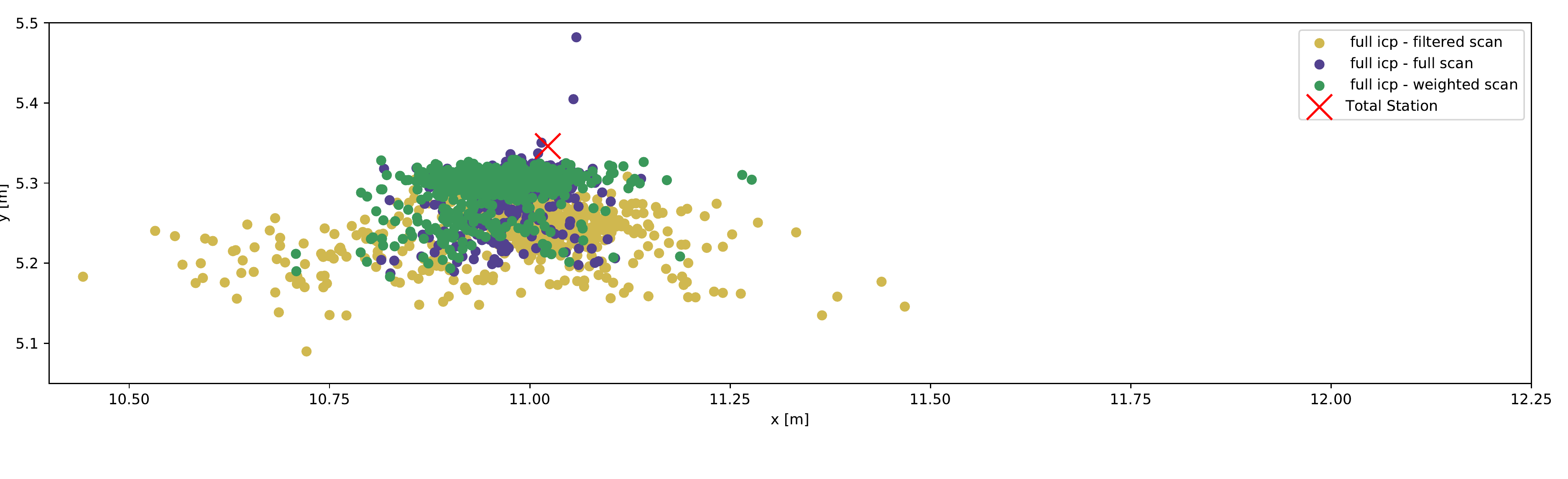}
    \end{subfigure}
    \begin{subfigure}[b]{\linewidth}
    \includegraphics[width=\linewidth, trim=0 380 0 410, clip]{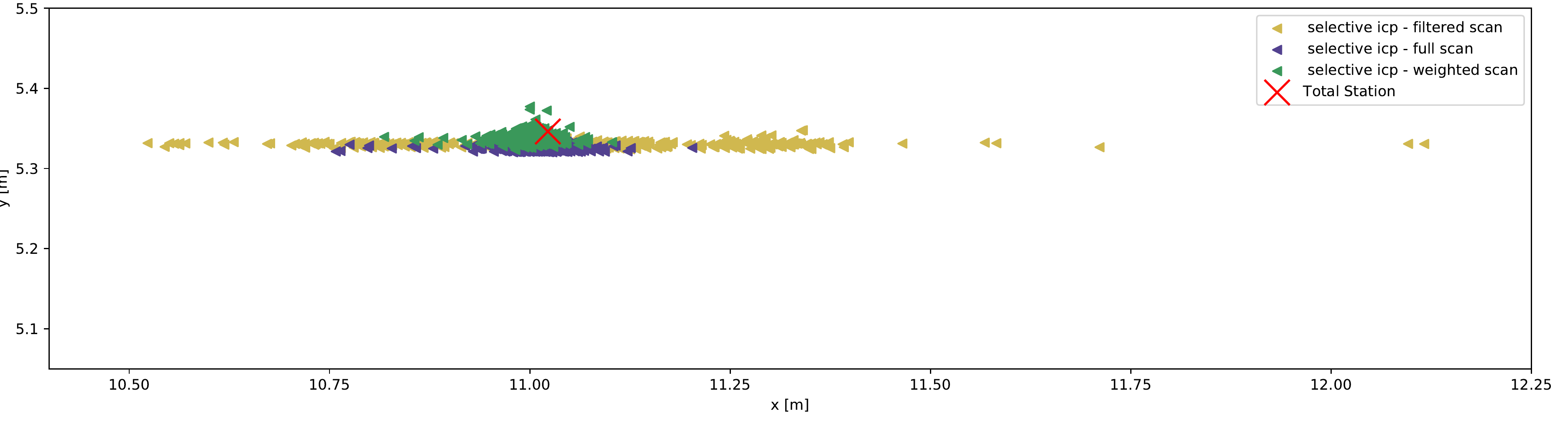}
    \end{subfigure}
    \caption{Topview on position C of the distribution of localization outputs for the stationary robot. All points are transformed to the estimated position of the tracking prism. The ground-truth location is given by the output of the total station for the tracked prism.}
    \label{fig:topview_c}
\end{figure*}

\begin{table*}
    \centering
\begin{tabular}{llrrrrrr}
\toprule
\multicolumn{2}{c}{Method}& \multicolumn{2}{c}{Pos. Repeatability} &  \multicolumn{2}{c}{Rot. Repeatabilty} & Accuracy &  Failure Rate \\
icp & scan & max eigenval.  & trace & max eigenval. & trace & rmse [mm] & [\%]\\
\midrule
full & full &        22.3 &       25.9 &         \textbf{2.7} &        \textbf{8.0} &       390 &           0.0 \\
full & filtered &         \textbf{1.5} &        \textbf{2.5} &         2.9 &        8.6 &       290 &           4.9 \\
full & weighted &        10.1 &       12.7 &         3.0 &        8.9 &       354 &           8.2 \\
selective & full &       121.1 &      127.2 &         5.3 &       15.9 &       452 &          50.3 \\
selective & filtered &        31.9 &       32.2 &         5.1 &       15.2 &       \textbf{232} &          44.1 \\
selective & weighted &       330.5 &      338.6 &        14.1 &       42.4 &       627 &          76.5 \\
\bottomrule
\end{tabular}

\caption{Stationary Localization Study Pos A}
\label{tab:stationary_a}
\end{table*}

\begin{table*}[]
    \centering
\begin{tabular}{llrrrrrr}
\toprule
\multicolumn{2}{c}{Method}& \multicolumn{2}{c}{Pos. Repeatability} &  \multicolumn{2}{c}{Rot. Repeatabilty} & Accuracy &  Failure Rate \\
icp & scan & max eigenval.  & trace & max eigenval. & trace & rmse [mm] & [\%]\\
\midrule
full & full &         3.3 &        5.9 &         \textbf{3.2} &        \textbf{9.6} &       222 &           0.0 \\
full & filtered &         \textbf{1.4} &        \textbf{2.3} &         3.4 &       10.3 &       342 &           5.3 \\
full & weighted &         3.8 &        5.6 &         3.6 &       10.7 &       230 &           8.1 \\
selective & full &         7.6 &        8.0 &         3.4 &       10.3 &       328 &           4.8 \\
selective & filtered &         2.6 &        2.7 &         4.2 &       12.6 &        \textbf{68} &          21.3 \\
selective & weighted &        17.1 &       18.4 &         7.0 &       21.0 &       244 &          52.6 \\
\bottomrule
\end{tabular}
    \caption{Stationary Localization Study Pos B}
    \label{tab:stationary_b}
\end{table*}

\begin{table*}[]
    \centering
\begin{tabular}{llrrrrrr}
\toprule
\multicolumn{2}{c}{Method}& \multicolumn{2}{c}{Pos. Repeatability} &  \multicolumn{2}{c}{Rot. Repeatabilty} & Accuracy &  Failure Rate \\
icp & scan & max eigenval.  & trace & max eigenval. & trace & rmse [mm] & [\%]\\
\midrule
full & full &         1.8 &        2.9 &         \textbf{4.3} &       \textbf{12.9} &        76 &           0.0 \\
full & filtered &        14.8 &       17.1 &         4.4 &       13.3 &       153 &           3.7 \\
full & weighted &         4.0 &        5.2 &         4.6 &       13.7 &       103 &           6.4 \\
selective & full &         \textbf{0.9} &        \textbf{1.0} &         \textbf{4.3} &       13.0 &        65 &           0.3 \\
selective & filtered &        28.7 &       28.9 &         5.9 &       17.6 &       154 &          25.6 \\
selective & weighted &         1.0 &        1.5 &         4.8 &       14.4 &        \textbf{52} &           9.8 \\
\bottomrule
\end{tabular}
\caption{Stationary Localization Study Pos C}
    \label{tab:stationary_c}
\end{table*}

\section{Discussion}
\label{sec:discussion}
From the results we see that in general, selective localization against reference walls constraints the localization very well in two dimensions. The trace and maximum eigenvalue for selective localization are very close, whereas there is uncertainty in more than one direction for classic ICP against the full building model. The same effect is illustrated in figure~\ref{fig:topview_c}. This raises the question why the localization shows such a high variance in lateral direction in all three locations. A more detailed evaluation of the test site shown in figure~\ref{fig:building_plan} reveals that most of the walls in the building model are parallel to the $x$ axis of the model and much fewer structures are parallel to the $y$ axis. For all 3 locations, however, the lateral localization depends on the alignment of surfaces parallel to the $y$ axis. The biggest available structure in this direction, which is also the reference wall for location C, is obstructed by a van parked in front of the wall. Therefore, for all three locations, the observable part of the lateral reference structure is small compared to both other directions and a higher localization uncertainty in $x$ direction had to be expected.

Our experiments further show that semantic filtering is more important for the selective ICP method. Intuitively, because the selective ICP aligns the scan only to a few reference surfaces, it is more vulnerable to obstructions of these surfaces. The full ICP alignment is also worse in presence of clutter, as shown in table~\ref{tab:clutter}, but our experiments show in two of three locations a better accuracy for the full scan than with selective ICP without semantic information. 

In all locations, the highest accuracy is achieved with a combination of semantic filtering of the LiDAR scan and selective localization to reference surfaces. However, we could not find a single combination of methods that always worked. We found the performance of the semantic filtering method to be highly dependent on the location and the respective performance of the density estimation network. As we already describe in section \ref{subs:method_semantic}, semantic classifiction in construction environments is difficult due to the lack of labelled data. The density estimation network from~\cite{Marchal2019-dv} showed reasonable performance for our test location, but still partially filtered out background structure that was too different from the training domain. In particular in location C, binary filtering of the pointcloud removed nearly all points on the lateral reference wall, resulting in high uncertainty and failure rate. We conclude that for precise localization of robots on construction sites outlier filtering and selection of high quality measurements are key aspects. While learning-based solutions show promising characteristics in this regard, we expect more domain-specific training datasets to further boost their performance.

\section{Conclusion}
\label{sec:conclusion}
In this work, we present a mobile on-board localization system for construction robots. In our experiments, we show that building deviations and clutter deteriorate accuracy of traditional registration methods, and find that a combination of semantic filtering of LiDAR scans and selective localization to reference walls yields essential accuracy gains. Our findings show a need for semantic datasets closer to the construction domain and a need for more research to further close the accuracy gap between external reference systems and on-board sensing.

\section*{Acknowledgement}
We would like to thank Selen Ercan for creating the 3D building models out of the existing plans. We would also like to thank Francesco Sarno for working with us on initial segmentation experiments.

\bibliographystyle{IEEEtran}
\bibliography{paperpile}
\end{document}